\documentclass[letterpaper, 10 pt,conference]{ieeeconf}  

\IEEEoverridecommandlockouts                              

\usepackage{amsmath,amssymb,amsfonts}
\usepackage{bm}
\usepackage{algorithm}
\usepackage[noend]{algpseudocode}
\usepackage{graphicx}
\usepackage{xcolor}
\usepackage[usenames,dvipsnames]{pstricks}
\usepackage{epsfig}
\usepackage{amssymb}
\usepackage{hyperref}
\usepackage{subfigure}
\usepackage{amsmath}
\usepackage{eso-pic}

\graphicspath{{src/}}

\usepackage{todonotes}

\newcommand{\valcahnge}[2]{\textcolor{blue}{} \textcolor{black}{ #2}}

\title{\LARGE \bf Maximally manipulable vision-based motion planning for robotic rough-cutting on arbitrarily shaped surfaces
}
\author{T. Pardi
$^{1,2}$, V. Ortenzi$^{1,2}$, C. Fairbairn$^{1,4}$, T. Pipe$^{1,5}$, A. M. Ghalamzan E.$^{1,3}$, and R. Stolkin$^{1,2}$.
\thanks{
$^1$ UK National Centre for Nuclear Robotics.
$^2$Extreme Robotics Lab (ERL) at the University of Birmingham, UK. \{txp754, v.ortenzi and R.Stolkin\}@bham.ac.uk. %
$^3$School of Computer Science, University of Lincoln, UK, aghalamzanesfahani@lincoln.ac.uk 
$^4$National Nuclear Laboratory, UK.
$^5$Bristol Robotics Lab (BRL), University of West England.
Tommaso Pardi is supported by a doctoral bursary of the UK Nuclear Decommissioning Authority.
Rustam Stolkin is supported by a Royal Society Industry Fellowship. This work forms part of the UK National Centre for Nuclear Robotics initiative, part-funded by EPSRC EP/R02572X/1. The work was also partially supported by Faraday ReLiB, EPSRC EP/P017487/1, EP/P01366X/1.
}
}

\begin{document}
\onecolumn
{\Huge  \bf IEEE Copyright Notice}
\newline
\newline
\textcopyright 2019 IEEE.  Personal use of this material is permitted.  Permission from IEEE must be obtained for all other uses, in any current or future media, including reprinting/republishing this material for advertising or promotional purposes, creating new collective works, for resale or redistribution to servers or lists, or reuse of any copyrighted component of this work in other works.
\newline
\newline
The work has been submitted to Robotics and Automation Letters in September 2019, with the ICRA 2020 option.

\twocolumn
\maketitle
\thispagestyle{empty}
\pagestyle{empty}

\begin{abstract}
This paper presents a method for constrained motion planning from vision, which enables a robot to move its end-effector over an observed surface, given start and destination points. The robot has no prior knowledge of the surface shape, but observes it from a noisy point-cloud camera. We consider the multi-objective optimisation problem of finding robot trajectories which maximise the robot's manipulability throughout the motion, while also minimising surface-distance travelled between the two points. This work has application in industrial problems of \textit{rough} robotic cutting, \textit{e.g.} demolition of legacy nuclear plant, where the cut path need not be precise as long as it achieves dismantling. We show how detours in the cut path can be leveraged, to increase the manipulability of the robot at all points along the path. This helps avoid singularities, while maximising the robot's capability to make small deviations during task execution, \textit{e.g.} compliantly responding to cutting forces via impedance control. We show how a sampling-based planner can be projected onto the Riemannian manifold of a curved surface, and extended to include a term which maximises manipulability. We present the results of empirical experiments, with both simulated and real robots, which are tasked with moving over a variety of different surface shapes. Our planner enables successful task completion, while avoiding singularities and ensuring significantly greater manipulability when compared against a conventional RRT* planner.

%
%
%

\end{abstract}

\section{Introduction}
\subsection{Background}
\begin{figure}[tb!]
    \includegraphics[trim=0.0cm 0.0cm 0cm 0.0cm, clip=true,scale=0.69,angle = 0]{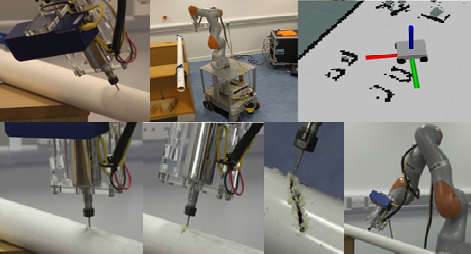}
    \caption{Proof-of-principle mobile manipulator robot, cutting a pipe with an axial rotary cutter. Robot developed by Bristol Robotics Lab, T. Pipe et al.$^{1,5}$. The kinematics of path-planning is similar to that for a laser cutter, in that the cutter axis must be maintained normal to the local surface curvature, and rotations of the robot around the tool axis are allowable. However, additional dynamics problems are engendered by forceful contact between robot and work-piece.
    \newline
    }
    \label{fig:BRLcutting}
    \vspace{-12pt}
\end{figure}

Robotic cutting actions engender an interesting problem of motion-planning for a serial arm under semi-closed chain constraints. The end-effector cutting tool is constrained to touch the cutting surface, thereby forming a closed chain at any given time step. However, the cutting surface can be regarded as a manifold upon which the end-effector has \textit{locally} two or three degrees of freedom to move (depending on whether rotations about the axis of the cutting tool are permitted, in addition to translations on the manifold). Note that we typically must align the cutting tool axis with the local surface normal, Fig. \ref{fig:BRLcutting}.

Here we are concerned with \textit{rough} cutting for problems such as robotic demolition in hazardous environments. In such applications, the exact cutting path is not important, as long as the robot successfully \textit{e.g.} cuts an object into two pieces, or cuts open a container so that its contents can be inspected.
\begin{figure}[t]
    \centering	  
    \includegraphics[trim= 0cm 0cm 2cm 3.5cm, clip=true,scale=1.0,angle = 0]{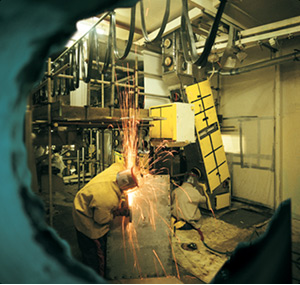}
    \hspace{\subfigtopskip}\hspace{\subfigtopskip}\hspace{\subfigtopskip}
  \caption{Nuclear decommissioning worker wearing air-fed plastic suit underneath heavy leather overcoat, and multiple layers of gloves, while using power tools to cut legacy nuclear plant contaminated by alpha-radiation emitting substances, such as plutonium dust. The leather coat protects the plastic suit from being punctured by hot sparks during cutting. Maximum 2hrs work per day is possible, due to extreme discomfort and heat exhaustion as the suit fogs and fills with sweat. Image courtesy of Sellafield Ltd.}
  \label{Fig:suitAndSuitWorker}
  \vspace{-12pt}
\end{figure}
A particular focus of our work is the use of robots for cleanup of legacy nuclear waste \cite{marturi2016towards, NCNR, H2020romans}, however other applications include \textit{e.g.} asbestos-contaminated buildings \cite{H2020bits2rec}, and emergency operations such as bomb-disposal, fire-fighting and disaster-response \cite{murphy2004human}.

The UK alone contains an estimated 4.9 million tonnes of legacy nuclear waste. Without significant advances in robotics, it is expected to require at least one million entries of human workers into radioactive zones, wearing cumbersome protective air-fed suits while cutting and dismantling contaminated structures Fig. \ref{Fig:suitAndSuitWorker}. For numerous higher radiation environments, no entry of humans is possible at all, and so there is no way to achieve decommissioning without remote manipulation technology.


A variety of tooling can be used for robotic cutting. Our team previously worked closely with the UK nuclear industry to achieve a world-first of autonomous vision-guided robotic laser cutting of contaminated metal inside a radioactive facility \cite{dodds2020radionuclide}. Lasers, and other non-contact methods such as water jet or plasma cutting, are convenient in that no contact forces are exerted, although close geometric surface following (with a few mm stand-off) must still be achieved.

In contrast, we are now experimenting with axial rotary cutting tools (similar to the cutter of a milling machine), Fig. \ref{fig:BRLcutting}. The kinematic path-planning constraints for such tools are similar to those for a laser, in that the cutter axis must be maintained normal to the local surface curvature, and rotations of the robot around the tool axis are allowable. However, with the rotary cutter, forceful interactions, between the robot and cut materials of uncertain properties, introduce significant perturbations. We would like the robot to \textit{have sufficient manipulability to provide capacity for responding compliantly to such perturbations}, while following a desired cutting path.

\subsection{Related work}
There is a large body of literature
on path-planning for tool paths in multi-axis CNC machining \cite{lin1996efficient}. This work is aimed at precision manufacturing, where it is essential that the cutting tool rigidly follows an exact path through the work-piece. In contrast, for our application, other factors such as safely avoiding singularity configurations, and allowing capacity for perturbations, are of much more importance than the exact cutting path. This is particularly the case in legacy nuclear sites, where material properties are significantly uncertain, partly due to lack of exact records after many decades, and also due to uncertain degradation of materials under radiation dose, corrosion and ageing.

A related problem is the use of serial robots for depositing paint, surface coatings or \textit{e.g.} applying heat treatments over the surface of components and products in manufacturing. Optimised end-effector (\textit{e.g.} paint-gun) trajectories should deposit an even thickness of paint, requiring a constant speed of the end-effector over the surface, constant stand-off distance, and constant distance between successive parallel ``mowing-the-lawn'' type traverses of the surface. However these constraints are less rigid than for CNC machining, and can be treated as fitness functions for numerically optimising the end-effector path. Originally such paint robots were hand-programmed ~\cite{Suh1991}. Later work plans ``mowing-the-lawn'' type traverses on a surface manifold modeled as \textit{e.g.} Bezier curves~\cite{Chen2017}. More recent work such as ~\cite{Freitas2017} models a work-piece as a set of surface patches, and uses route searching approaches such as Fast Marching to find end-effector paths which link all such patches together to achieve coverage.

However, unlike our work, these paint-planing methods predominantly focus on planning the route of an end-effector tool, without taking into account the corresponding inverse-kinematics and robot configurations. These methods also rely on explicit 3D CAD models of the work-piece as a-priori knowledge. They also tackle a problem where the tool path itself is of prime importance. Since they deal with known manufactured parts, it is acceptable to devote large computational resource and long run-times to optimise trajectories off-line prior to repeated execution on many identical objects.

In contrast, nuclear decommissioning or disaster response involves highly unstructured environments, Fig. \ref{Fig:suitAndSuitWorker}, and we must plan cuts in near-to-real time on unknown objects observed by noisy partial point-cloud views. Furthermore, we are not concerned about following an exact cutting path. Instead, it is important for us to consider inverse-kinematics throughout the planned motion, and we modify the cutting path itself to avoid singularity configurations in the arm, and to improve robustness to perturbations by maximising manipulability throughout the motion.

There has been comparatively little work done by the robotics research community on cutting. Recent literature on robotic tool use includes \textit{e.g.} \cite{peternel2016towards}, in which a co-bot learns to assist a human with a backwards and forwards sawing motion. However, this work does not consider actually planning the cut. The work of \cite{toussaint2018differentiable} explores intelligent tool use more broadly, in the sense that the robot should sequentially grasp and use tools such as hooks and sticks, to push or drag objects on a table, during physical puzzle solving (using implements to reach and move distant objects). The main contribution is a hybrid planning approach, that combines high-level logical planning of sequences of action-primitives (\textit{e.g.} ``push'' or ``grasp''), with low-level motion planning of each action, exploiting physics simulators to assist with prediction. Related work, combining task-level logical planning with low-level motion planning for manipulation, includes \cite{dearden2014manipulation}, but this is focused on grasping with pick-and-place tasks.

There is now a large body of literature, and well established set of robust methods for robot path-planning with obstacle avoidance. Early work by Khatib modelled obstacles as artificial potential fields, and optimised collision-free paths by descending an energy gradient. More recent methods, for path-planning by gradient descent of cost functions, include the well-known CHOMP \cite{ratliff2009chomp} and stochastic variant STOMP \cite{kalakrishnan2011stomp}. It is now common to use sampling-based methods, such as PRM \cite{amato1996randomized} and RRT \cite{Lavalle98RRT}, to generate a net of points which can be explored to find collision-free paths. Later variants include RRT* \cite{Karaman2011, GaSrBa14} with improved convergence properties.

In this paper, we are not concerned with the typical formulation of the path-planning problem, \textit{i.e.} finding a shortest collision-free path. Instead, we wish to plan non-shortest paths which are optimal in other ways with respect to additional information and considerations (maximising manipulability under constraints). Various authors have sought to augment the classical path-planning approaches by incorporating modified cost functions, based on additional kinds of information, to induce useful additional robotic behaviours.

A cost-based optimisation approach was proposed in \cite{kruger2007optimal}, to enable an Autonomous Underwater Vehicle (AUV) to plan a path between specified start and destination locations. The method exploits computational ocean model forecasts of water current speeds and directions at different locations, depths and times. The robot plans large deviations in its route, to avoid adverse currents, while exploiting (\textit{i.e.} ``riding'' on) currents in useful directions to minimise energy expenditure during the journey. The route is optimised according to a multi-objective cost function, which includes separate terms for avoiding obstacles, minimising energy expenditure, minimising journey time and other considerations. Related work in the Unmanned Aerial Vehicle (UAV) literature also considers environmental factors, \textit{e.g.} \cite{Nikolos2003}, and planning on non-flat surfaces with constraints is considered in the ground-vehicle \cite{Nikolos2003} and legged locomotion~\cite{Lin2018,Kanoulas2018} literature. The estimation of kinematic constraints from contacts with surfaces is also analysed \textit{e.g.} in \cite{Ortenzi2016}.

Unmodified conventional path planning algorithms are not immediately useful for computing a cutting path on an object surface suitable for a robotic manipulator. An end-effector (cutting tool) path computed with these approaches may be out of the reachable workspace of the manipulator or may pass through singular configurations of the robot. Furthermore, in demolition rough-cutting scenarios, we are dealing with a highly unstructured environment, in which a-priori 3D models for the cut object are typically unavailable.

At any particular instant during cutting, a serial arm is constrained to form a closed chain with the cut surface. There is a body of prior literature that explores end-effector constraints during the path generation. Combining closed-chain constraints with sampling-based planning (PRM) was explored in \cite{han2001kinematics} for motion of parallel manipulators. In \cite{berenson2011task}, end-effector constraints are characterised as ``Task Space Regions'' (TSRs), which can encode \textit{e.g.} the constrained trajectory of a robot hand when opening a door. Planning under such constraints is handled by sampling from an appropriate TSR. Unlike our work, \cite{berenson2011task} explores tasks where a variety of destination poses are possible, \textit{e.g.} bi-manual constraints on a pair of arms that must grasp a box while placing it anywhere on a table. Furthermore, manipulability during task execution is not explored in \cite{berenson2011task}, and complete a-priori 3D knowledge of the scene is assumed.

In \cite{Jaillet2008} and \cite{5477164}, a transition-based RRT algorithm driven by a cost based on mechanical work generates a collision-free path between points. Similarly, a heuristically biased RRT is proposed in \cite{UrmsonS03} to guide the search on the tree. Inverse kinematics guide the search in the implementation of RRT in \cite{BertramKDA06}, while the manipulability of the robot is exploited to bias the sampling process in \cite{Lavrovsky2003}. A higher manipulability over the path might improve the likelihood of succeeding in performing the task as it has been discussed in \cite{ghalamzan2016task} for grasp planning.

\subsection{Contributions of this paper}
%
%
The contribution of this paper is twofold: (i) RRT* is adapted to obtain a path with higher robot manipulability combining end-effector path-planning with kinematic considerations; (ii) we use logarithmic (and exponential) mapping to generate samples on the raw point cloud of the object to cut, thus making our approach model-free in terms of object models. 
Our experimental results show that the robot successfully follows the path obtained by our approach whereas the robot may fail to follow a path obtained by a naive RRT*.
Although our work is motivated by sort and segregation of nuclear waste by industrial manipulators, the proposed approach can be readily used in many other domains, \textit{e.g.}, a humanoid robot can use the computed cutting path to peel an orange with a knife.

 \begin{figure}[tb!]
    \centering
    \subfigure[][]{ \includegraphics[trim=3cm 2cm 3cm 1.75cm, clip=true,scale=.2,angle = 0]{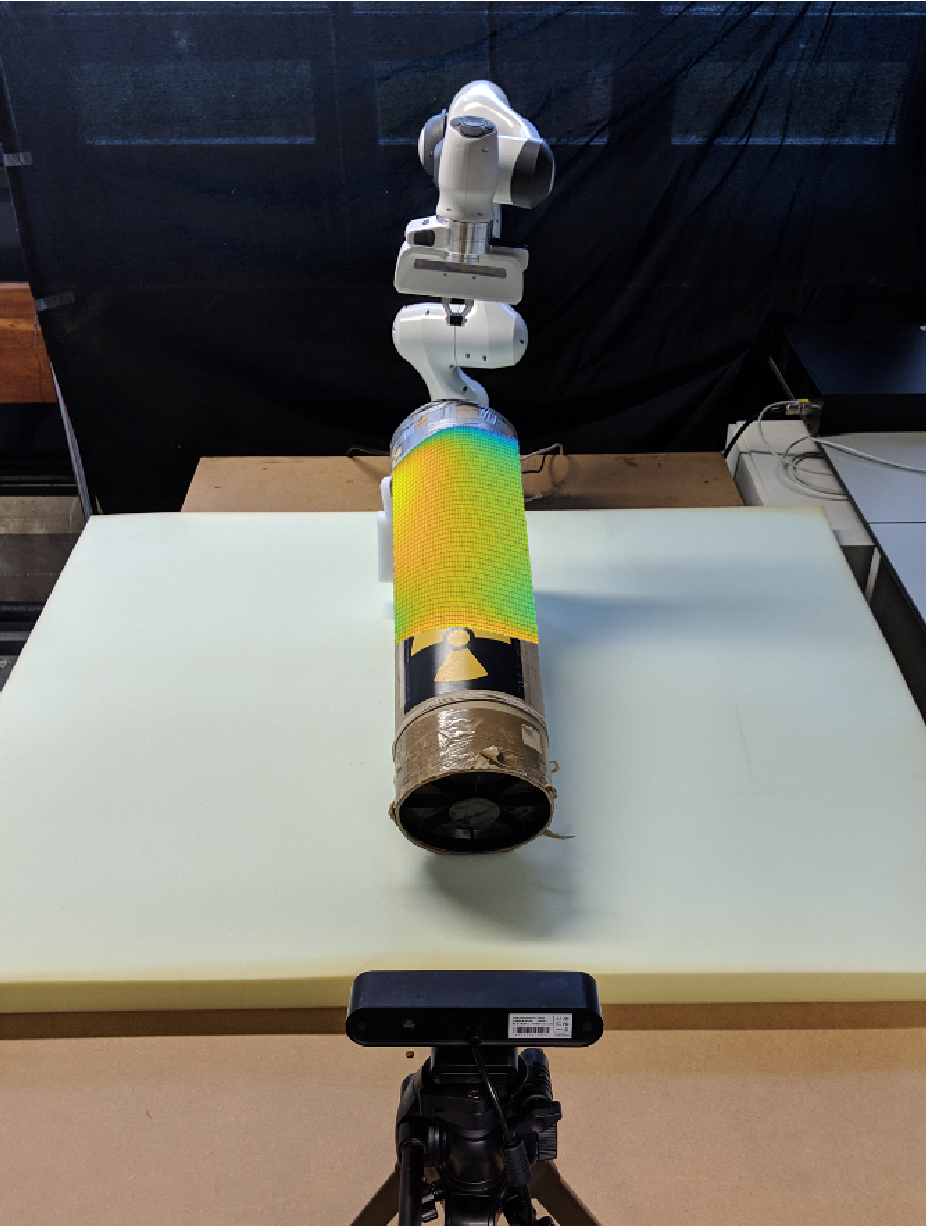}
     \label{fig::franka_setup}}
     \subfigure[][]{\includegraphics[trim=8cm 9.7cm 8cm 9.5cm, clip=true,scale=.46,angle = 0]{setup_experiment_nuclear_waste_overlapped}
     \label{fig::man_map}}
    \vspace{0pt}
  \caption{An experimental robotic cutting setup (Fig.~\ref{fig::franka_setup}). A 3-D camera (positioned in front of the robot at 2.8 [m] distance facing the robot) captures the point cloud of the object surface. Our approach computes a cutting path with given initial and end points. This path is suitable for the robot kinematics as our algorithm accounts for a manipulability index.  Fig.~\ref{fig::man_map} shows the manipulability of the robot: each point is coloured based on the manipulability corresponding to the configuration the robot is in when its end effector touches such point. Points with lower manipulability are shown in blue; points with higher manipulability are represented in yellow and red.
  }
  \label{Fig::exp_setup_object}%
\vspace{-12pt}
\end{figure}



%
%
%
\section{Problem formulation}
\label{sec:prob_def}
We use RRT* to generate a path from point A to point B.
Moreover, we incorporate \emph{Riemannian manifold mapping} into our approach to generate samples that lie on the point cloud of the object.
We also include a cost function based on the manipulability of the robot into RRT*.
These modifications to the naive RRT* guarantee that the computed path (i) connects the start point (A) and the end point (B) possibly specified by a user, (ii) lies on the object surface and (iii) is feasible for the manipulator.
%
%
As such, our approach is called ``RRT*-RMM" which stands for RRT* with added Riemannian Manifold mapping and added Manipulability cost. 
\subsubsection*{Rapidly-exploring Random Tree*}
\label{sec:rrtstar}
Rapidly-exploring Random Tree (RRT) is one of the most common sampling-based path planners, \cite{Lavalle98RRT}. 
The basic idea behind RRT is to sample points within a region of interest and add them in a tree structure based on a distance metric. 
Every iteration, the algorithm generates a new point based on some motion constraints and, then, connects it to the closest node in the tree. 
RRT* is an extension to the classical RRT proposed in \cite{Karaman2011}, which allows the re-evaluation of nodes already in the tree when a new point is available. 
This procedure is usually referred to as \textit{rewiring}. 
During the rewiring, the algorithm selects the neighbourhood of a point (points in the tree within a range distance to the point) and evaluates whether these nodes improve their value passing through the new available point. 
This process provides RRT with better convergence to a solution and the
solution converges to the shortest path as the number of samples goes to $\infty$.

%
%
\subsubsection*{Manipulability} Let $\bm{q} \in \mathbb{R}^n$ represent the robot configuration where $n$ is the number of degrees of freedom (dof) of the robot.
Given a specific $\bm{q}$, position and orientation of every point of the robot are uniquely defined (forward kinematics).
This mapping, $\bm{f_r}$, is commonly expressed as
\begin{equation}
    \bm{r = f_r(q)},
    \label{eq::directmapping}
\end{equation}
where $\bm{r} \in \mathbb{R}^m$ is the position and/or the orientation of a point of interest of the robot in the Cartesian space and $m$ is the dimension of this representation (\textit{e.g.}, $m = 3$ for 3D position, or $m = 6$ for 3D position and orientation). Differential kinematics are defined using the robot Jacobian $\bm{J(q)}$ as
\begin{equation}
    \bm{\dot{r}= J(q) \dot{q}}
\end{equation}
and relate velocities in the configuration space to velocities in the Cartesian space\footnote{Since the Jacobian matrix always depends on the configuration $\bm{q}$, we drop the dependence on $\bm{q}$, and in the following we write $\bm{J(q)}$ as $\bm{J}$.}. If we constrain the norm of the configuration velocities to be unitary, the configuration lies on the unitary sphere $\mathbb{S}^1$
\begin{equation}
    \bm{|\dot{q}| = \dot{q}^T \dot{q} = \dot{r}^T J^{\dagger T} J^{\dagger} \dot{r} = \dot{r}^T \Gamma^\dagger \dot{r}  = 1}
\end{equation}
where $\bm{^\dagger}$ is the inverse matrix when $\bm{J}$ is square or the pseudo-inverse matrix when it is not.

Previous work leverages manipulability to yield optimal manipulation movements for planning a suitable grasping pose~\cite{ghalamzan2017human}. 
We would also like to optimise the manipulation capability for robotic cutting.
The conventional measure of manipulability \cite{Yoshikawa1985} is defined as
\begin{equation}
\label{eq:manYoshi}
    w(\bm{q}) = \sqrt{det(\bm{J^TJ})}  = \sqrt{\lambda_1 \lambda_2...\lambda_n},
\end{equation}
where $\lambda_i$ are the eigenvalues of $\bm{\Gamma^T}$.
This index provides a value that is proportional to the volume of the manipulability ellipsoid, and it does not require a long computational time.
\subsubsection*{Riemannian Manifold}
Computing distances between points is not straightforward on curved surfaces. 
By definition, a manifold is an n-dimension topological space that approximates the Euclidean space in the neighbourhood of any of its points. 
Furthermore, a Riemannian Manifold 
is defined as a smooth manifold $\mathcal{M}$ equipped with an inner product $g$ on the tangent space $T_p\mathcal{M}$ of each point $\bm{p}\in\mathcal{M}$, which changes smoothly from point to point and its vector spaces are differentiable. The family of inner products on the manifold is called Riemannian metrics.
\begin{figure}[tb!]
    \includegraphics[trim=-4cm 0.0cm 0cm 0.0cm, clip=true,scale=.2,angle = 0]{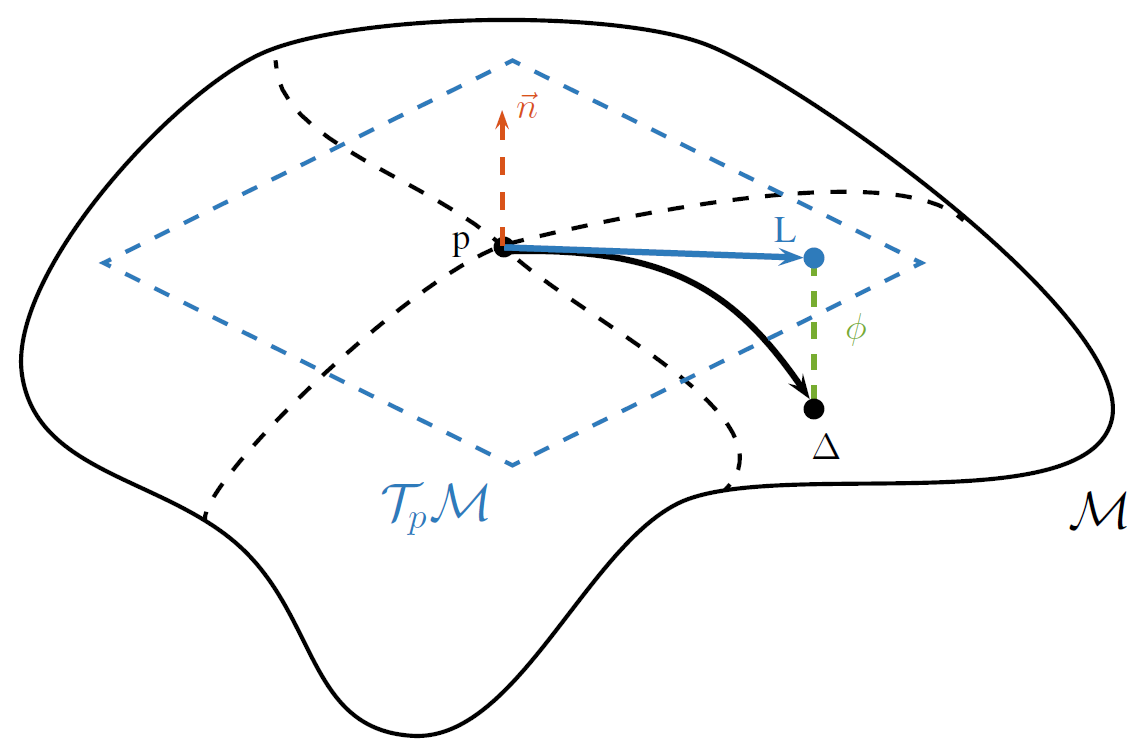}
    \caption{The figure shows an example of the exponential map, $\phi$, which projects the point $L\in T_p\mathcal{M}$, where $T_p\mathcal{M}$ is the tangent space to the point $\bm{p} \in \mathcal{M}$, to the manifold $\mathcal{M}$.}
    \label{fig:exp_log_maps}
    \vspace{-12pt}
\end{figure}
Let $\mathcal{M}$ be a manifold, $\bm{p}$ a point on $\mathcal{M}$, and let $\bm{v}\in T_p\mathcal{M}$ be a tangential vector to the manifold at point $\bm{p}$. Then, there is a unique geodesic $\gamma_v(s)$, with $s\in[0,1]$, such that $\gamma_v(0) = \bm{p}$ is satisfied and with initial tangent vector $\dot{\gamma}_v(0) = \bm{v}$. 
Under these assumptions, we can interpret the vector $\bm{v}$ as the linear velocity at the point $\bm{p}$ along the trajectory $\gamma_v(s)$.
Therefore, in a neighbourhood of $\bm{p}$, it is possible to define a map, called \textit{exponential map} (Fig. \ref{fig:exp_log_maps}), such that
\begin{equation}
    \phi = \exp_p(L) = \gamma_v(1).
    \label{eqexp}
\end{equation}
This map $\phi: T_p\mathcal{M} \mapsto \mathcal{M}$ projects every point $L$ in $T_p\mathcal{M}$ onto the manifold $\mathcal{M}$. 
Thanks to this \textit{exponential map}, we are able to generalise the concept of straight line in Euclidean space to curved surfaces, and the new metric that measures the distance between two points onto the manifold is called \textit{geodesic}.
Conversely, we can define a function which moves elements from the manifold into the tangent space of $\bm{p}$. This map is usually called \textit{logarithmic map}, and it is defined by
\begin{equation}
    \phi^{-1} = \log_p(\Delta)
    \label{eqlog}
\end{equation}
where $\phi^{-1}: \mathcal{M}\mapsto T_p\mathcal{M}$, and $\Delta$ is a point on the manifold $\mathcal{M}$, as in Fig. \ref{fig:exp_log_maps}.
\newline
\subsubsection*{Proposed method RRT*-RMM} Algorithm \ref{alg:rrtstarM} shows the pseudocode of our proposed approach.
Our RRT* method randomly selects a point on the point cloud and finds the closest point, denoted by $x_{nearest}$, in the tree (lines 3 and 4 in Alg. \ref{alg:rrtstarM}).  
We then compute a tangent plane to the closest point (line 5 in Alg. \ref{alg:rrtstarM}) on which the random point is projected, denoted by $x_{r-p}$, (line 6 in Alg. \ref{alg:rrtstarM}). 
We obtain a new point, denoted by $x_{new-p}$, by linear combination of the closest point and the projected point.
\begin{equation}
    x_{new-p} = x_{nearest} + \beta (x_{r-p} - x_{nearest})
    \label{lincomb}
\end{equation}
To satisfy the assumption required for logarithmic and exponential mapping between a Riemannian and Euclidian manifold in eqs.~\eqref{eqexp} and \eqref{eqlog}, we choose a small step size at this phase, i.e. $\beta\ll 1$. 
As per eq.~\eqref{lincomb}, we assure a small distance between closest point and new point, \textit{i.e.}, $x_{nearest}$ and $x_{new-p}$ are very close. 
These three points ($x_{nearest}$, $x_{new-p}$ and $x_{r-p}$) lie on the tangent plane (as shown in Fig.~\ref{fig:exp_log_maps}).
%
 \begin{algorithm}[tb!]
\caption{RRT*-Riemannian-mapping-manipulability}\label{alg:rrtstarM}
\begin{algorithmic}[1]
\State T.init($x_{init}$)
\For {$k=1~to~K$} 
\State $x_{rand} \gets RandomState()$
\State $x_{nearest} \gets Nearest(x_{rand})$
\State $TpM \gets computeTangentPlane(x_{nearest})$
\State $x_{r-p} \gets projectonTangentPlane(x_{rand}, TpM)$
\State $x_{new-p} \gets getNewPointontonTpM(x_{r-p}, TpM)$
\State $x_{new} \gets expRiemannianMap(x_{new-p})$ 
\If {$ObstacleFree(x_{new})$}
\State $X_{near} \gets Near(x_{new})$
\State $q_{new} \gets Ik(x_{new})$
\For{$k=1~to~N$}
\State $C_d \gets computeDistance(x_{new}, X_{near}(k))$
\State $C_{\bm{M}} \gets computeMan(q_{new}, X_{near}(k))$
\State $C(k) = (1-\alpha)C_d + \alpha C_{\bm{M}}$
\EndFor
\State $x_{p} \gets selectMinimumElement(C, X_{near})$
\State T.add\_vertex($x_{new}, x_{parent}$)
\State T.add\_edge($x_{parent}, x_{new}$)
\State T.rewire($X_{near}, x_{new}, q_{new})$
\EndIf
\EndFor
\State \Return T
\end{algorithmic}
\end{algorithm}
 \begin{figure*}[tb!]
    \centering
     \subfigure[][]{\includegraphics[trim=0cm 0cm 0cm 0cm, clip=true,scale=1.6,angle = 0]{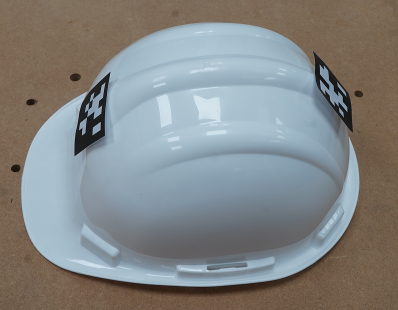}
     \label{fig::helmet_real}}
     \subfigure[][]{ \includegraphics[trim=1cm 0cm 3cm 2.0cm, clip=true,scale=.19,angle = 0]{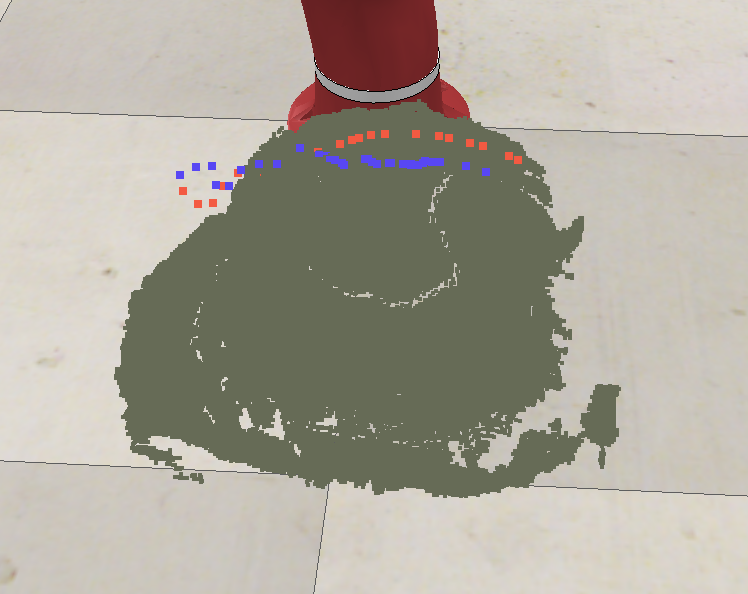}
     \label{fig::helmet_vrep}}
     \subfigure[][]{ \includegraphics[trim=0cm 0cm 0cm 0cm, clip=true,scale=.25,angle = 0]{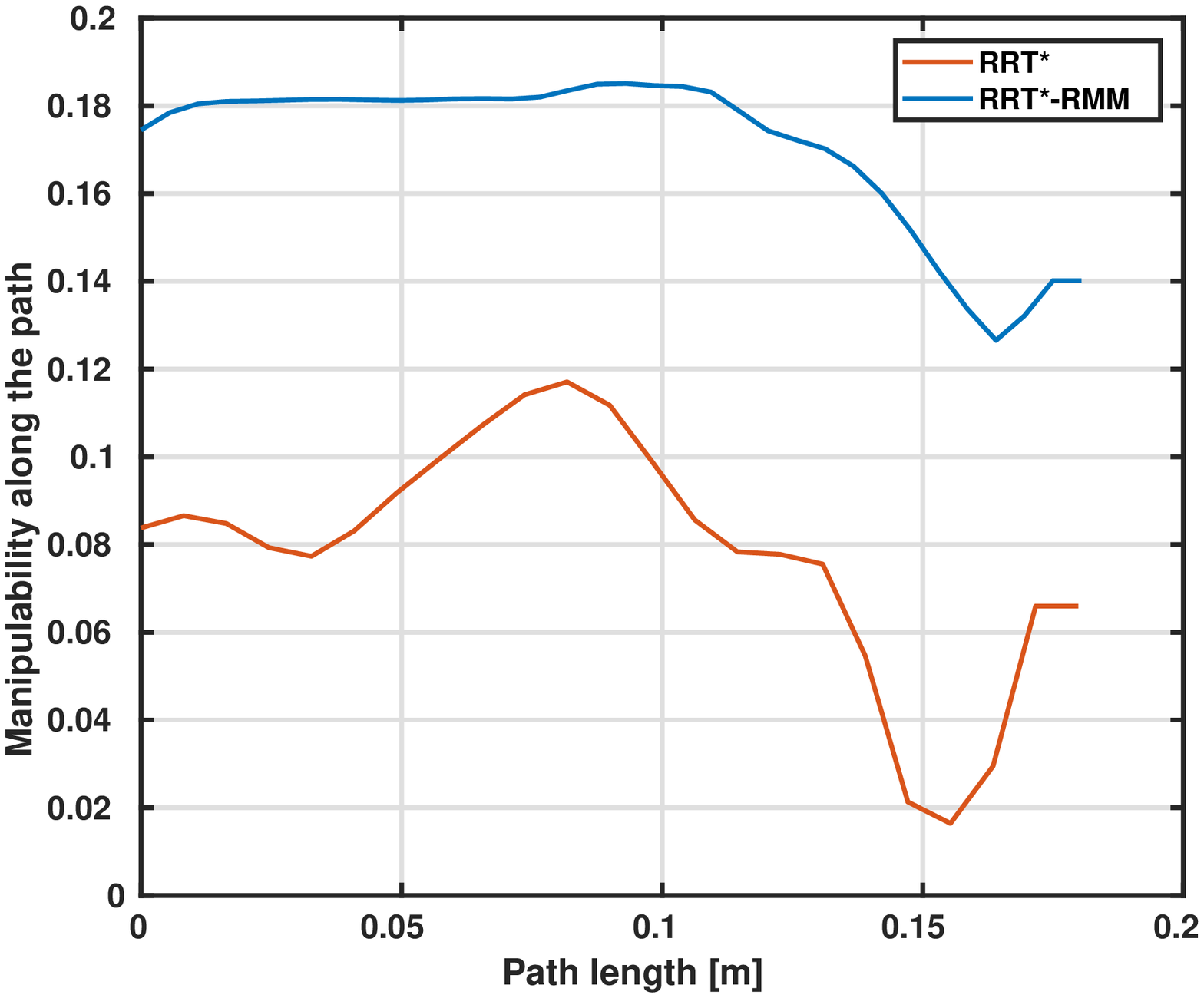}
     \label{fig::helmet_man}}
    \vspace{0pt}
  \caption{This image shows the path proposed by the two algorithms for a path connecting the two points indicated with the markers. \ref{fig::helmet_vrep} shows the point cloud of the helmet captured by the camera in V-REP and the path proposed by RRT* and our proposed approach are shown in red and blue dotted lines, respectively. In \ref{fig::helmet_man}, the manipulability of both paths is shown.}
  \label{Fig::helmet}%
\vspace{-12pt}
\end{figure*}
We introduce a modified cost (lines 14, 15 and 16 in Alg. \ref{alg:rrtstarM}) to be used in our RRT*, which is the sum of a "manipulability" cost along the path scaled by the number of elements in the path and a cost of "distance" from the starting point, as per eq.~\ref{eq:cost_total}. 

\begin{equation}
    C(\bm{p}) = (1-\alpha) C_d(\bm{p}, \bm{p_S}) + \alpha C_{\bm{M}}(\bm{q}_p) 
    \label{eq:cost_total}
\end{equation}
where $\bm{p} \in \mathbb{R}^3$ is the evaluated point, $\bm{p_S} \in \mathbb{R}^3$ is the starting point, $C_d$ is the distance cost, $C_{\bm{M}}$ is the manipulability cost accounting for the robot configuration ${\bm{q}_p}$ which is obtained using the robot's inverse kinematics at point $\bm{p}$, \textit{i.e.}, $\bm{q}_p = \bm{f_r}^{-1}(\bm{p})$. 
The coefficient $\alpha\in[0,1]$ is a trade-off between the costs and weighs the two contributions appropriately. In other words, when $\alpha \rightarrow 0$, the proposed algorithm turns into the classical RRT*; while $\alpha \rightarrow 1$ puts all the importance on the manipulability, discarding any consideration on distance. Such parameter must be chosen based on domain knowledge. 
As per eq.~\eqref{eq:cost_d}, we compute the sum of all segments over the path to reach $\bm{p} = \bm{p}_{N_p}$ where $\bm{p}_1$ is the initial point in eq.~\eqref{eq:cost_total}.
\begin{equation}
    C_{d}(\bm{p}) = \sum_{n_p=1}^{N_p} g(\bm{p}_{n_p}, \bm{p}_{n_p-1})
    \label{eq:cost_d}
\end{equation}
where $N_p$ is the number of points visited in the tree until reaching $\bm{p}_{N_p}$ from $\bm{p}_1$ and $g(.)$ is the geodesic distance between two adjacent points in the path. 
Assuming adjacent points are very close this geodesic distance can be approximated by $g(\bm{p}_{n_p}, \bm{p}_{n_p-1}) = \|\bm{p}_{n_p} - \bm{p}_{n_p-1}\|$. 
The manipulability cost is also computed over the path, as per eq.~\eqref{eq:cost_m}.
\begin{equation}
    C_{\bm{M}}(\bm{q}) = \frac{1}{N_p} \sum^{N_p}_{n_p = 1} \frac{1}{w(\bm{q}_{n_p})}
    \label{eq:cost_m}
\end{equation}
where  $w$ is manipulability index presented in eq. \ref{eq:manYoshi}. We use the inverse of $w$ to make the cost suitable to sum with RRT* cost to be minimised. 
Minimising the manipulability cost is equivalent to maximising manipulability (in eq.~\eqref{eq:manYoshi}). 
%

To summarise, the computed path is the result of a trade-off between the minimum travelling distance between start and end point and the maximum manipulability of the robot while following that path. 
Also, we use the Riemannian manifold mapping described earlier to project a random generated point onto the object surface (point cloud). 

\section{Experimental results}
We use a Panda robot manufactured by Franka EMIKA for the real-world experiments\footnote{Experimental results are reported in the attached video.}.
We also provide some results with Sawyer in V-REP\footnote{ Although the algorithm needs the robot kinematics, our ROS implementation takes the robot URDF directly from the ROS server parameter. Therefore, we present some data collected with Panda and some others with Sawyer to show the robustness of our approach to changes of the kinematic chains.}. 
Both are 7-DOF robotic arms equipped with a standard parallel jaw gripper. 
An Orbbec Astra RGB-D camera scans the area in front of the robot (Fig. \ref{Fig::exp_setup_object}), and we remove points outside the robot's workspace as preprocessing filtering on the point cloud. 
The camera is calibrated with respect to the robot base frame. 
As such, we can express the point cloud captured by the camera in the robot base frame. 
Furthermore, we attach two markers to each object, as shown in Fig.~\ref{fig::helmet_real}. 
These markers represent the start point A and the end point B of the path and allow a human operator to select the initial and end points of the cut.
In this work, we do not cope with developing a controller for the robot to follow the trajectory. We, therefore, overestimate the point cloud to be able to move the robot throughout the path without jeopardising it.
The RGB-D camera takes the point cloud of the scene in front of the robot as input, and our algorithm computes a cutting path between point A and point B. 
\newline
Fig. \ref{Fig::exp_setup_object} shows the experimental setup with a cylindrical container emulating a nuclear waste barrel. 
A heat map overlaid on the surface of the object represents the manipulability corresponding to the configuration the robot is in when its end-effector is at the point on the object.
We use the standard inverse kinematics (IK) of the robot to compute the joint configurations and we use the same IK to move the robot.
We developed a full ROS package to compute the optimal cutting path. 
As the robot URDF can be loaded onto the parameter server and used by the algorithm, we can easily repeat our computation with any manipulator whose URDF is available. 

 \begin{figure}[tb!]
    \centering
    \includegraphics[trim=0cm 0cm 0cm 0cm, clip=true,scale=1.5,angle =0]{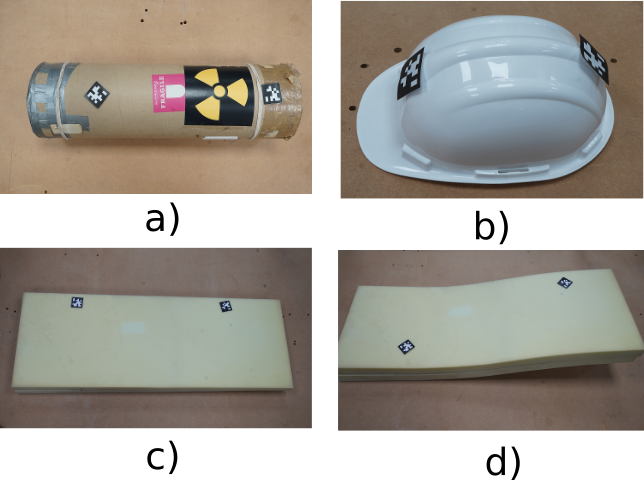}
  \caption{ Test objects used for testing our algorithm, (a) a barrel, (b) a curved object, (c) a helmet and (d) a flat object. The pictures also show the position of the markers.}
  \label{Fig::objects}%
\vspace{-12pt}
\end{figure}

 \begin{figure*}[tb!]
    \centering
    \hspace{-15pt}
     \subfigure[]{ \includegraphics[trim=0cm 0cm 0cm 1cm, clip=true,scale=.13,angle =0]{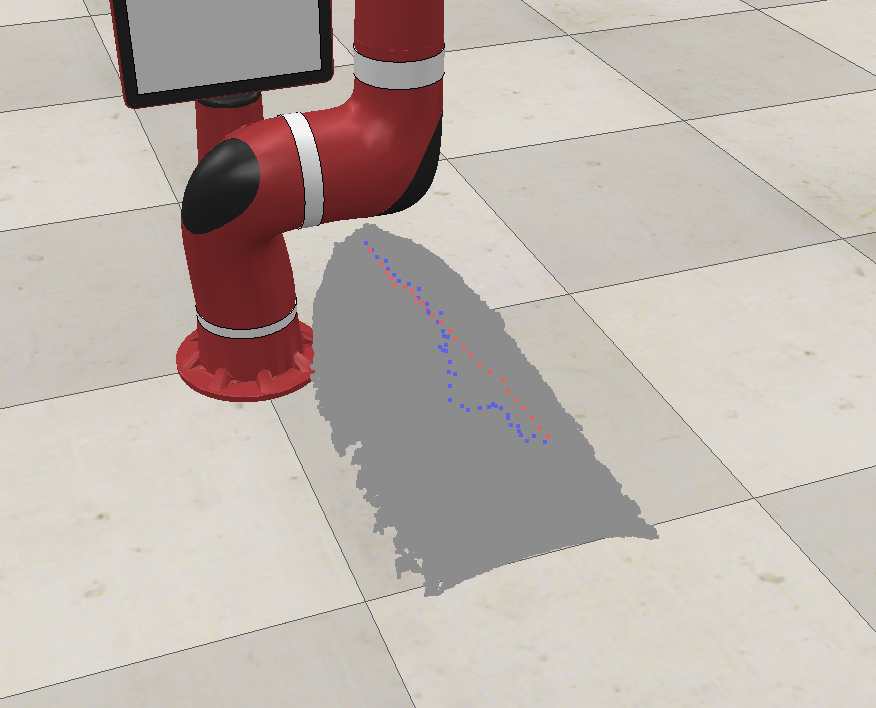}
     \hspace{\subfigtopskip}\hspace{\subfigbottomskip}
     \label{fig::barrel_vrep}}\hspace{-15pt}
     \subfigure[]{ \includegraphics[trim=0cm 0cm 0cm 0cm, clip=true,scale=.14,angle = 0]{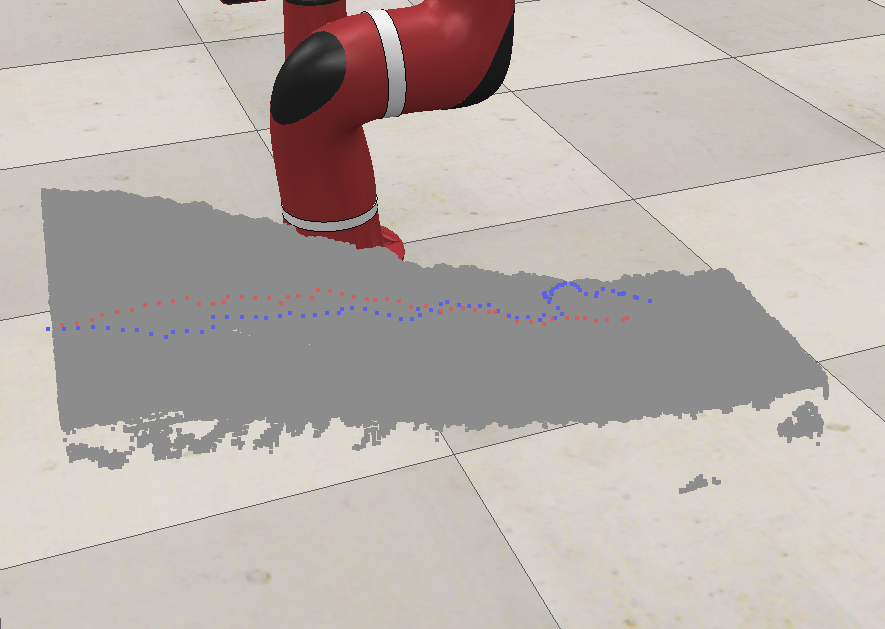}
     \hspace{\subfigtopskip}\hspace{\subfigbottomskip}
     \label{fig::cuboids_vrep}}\hspace{-15pt}
     \subfigure[]{ \includegraphics[trim=0cm 0cm 0cm 4cm, clip=true,scale=.14,angle = 0]{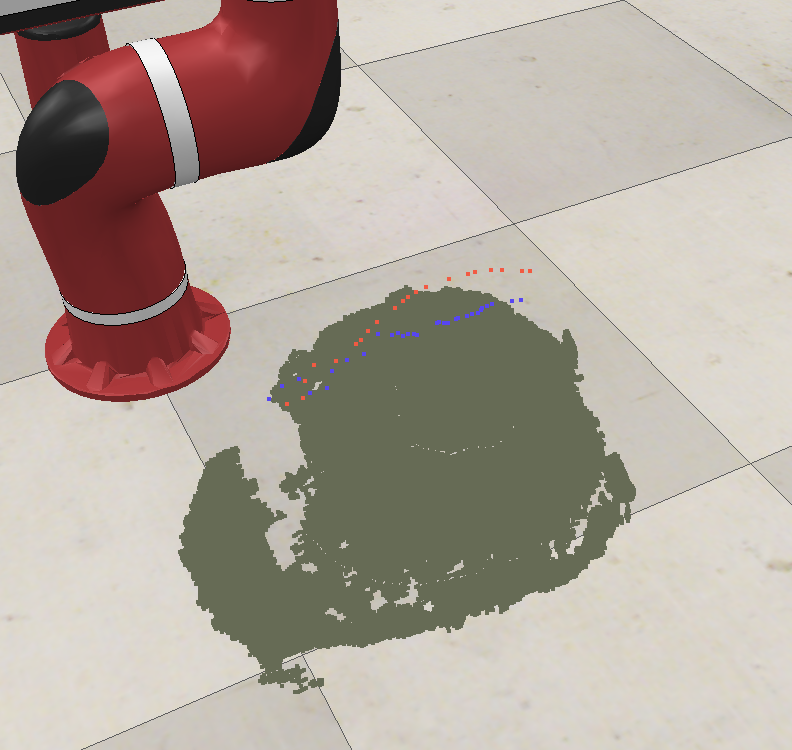}
     \hspace{\subfigtopskip}\hspace{\subfigbottomskip}
     \label{fig::helmet_vrep_1}}\hspace{-15pt}
     \subfigure[]{\includegraphics[trim=0cm 0cm 0cm 2cm, clip=true,scale=.12,angle = 0]{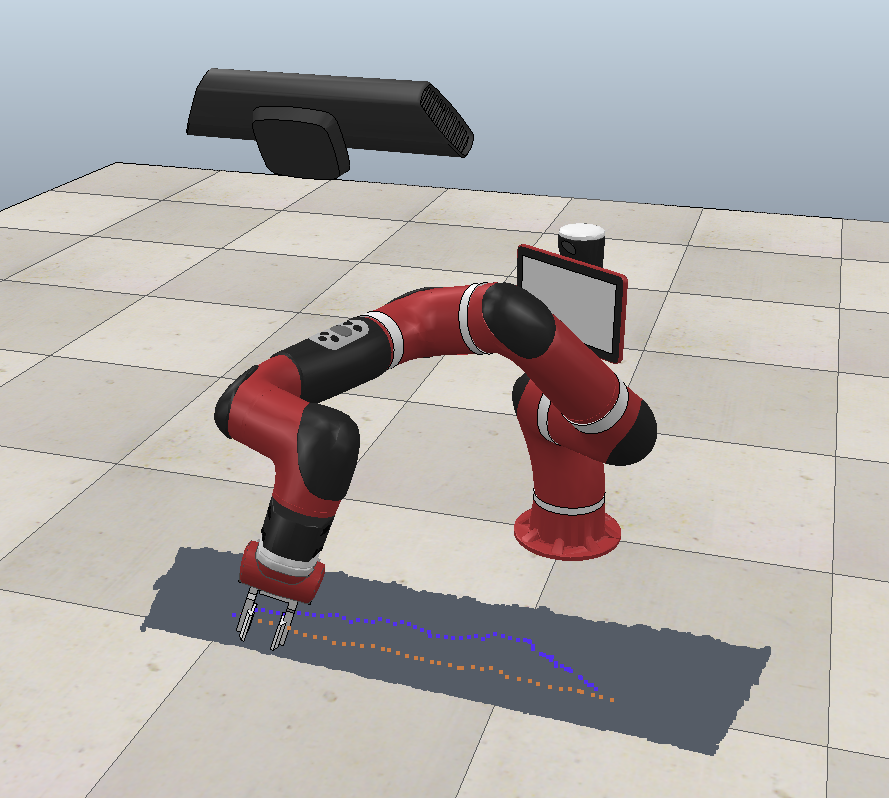}
     \label{fig::cuboid_vrep}}
     \hspace{-15pt}
    \vspace{0pt}  
    \caption{Paths proposed by RRT* and RRT*-RMM for the four objects, using initial and target positions shown in Fig. \ref{Fig::objects} using the markers. Fig. \ref{fig::cuboids_vrep} shows how RRT* proposes a path close a singularity for the robot (orange path) instead, the RRT*-RMM does a semicircle route to avoid it. In our experiments, we empirically selected $\alpha = 0.7$ to trade-off path's length and manipulability. }
  \label{Fig::vrep_path}%
\vspace{-1pt}
\end{figure*}
We used four objects (Fig.~\ref{Fig::objects}) to illustrate the effectiveness of our approach in generating a cutting path on different objects surfaces.
These objects are a barrel (cylindrical container), a curved object (made of foam), a safety helmet and a flat object (also made of foam). 
These objects represent typical objects that are to be cut in a nuclear environment.
Future work will include the deployment of the proposed algorithm to cut real-world object found in such environments.

Fig.~\ref{Fig::helmet} shows the helmet we used for our experiments along with the markers attached to the object.
 \begin{figure}[tb!]
    \includegraphics[trim=0cm 0cm 0cm 0cm, clip=true,scale=.4,angle = 0]{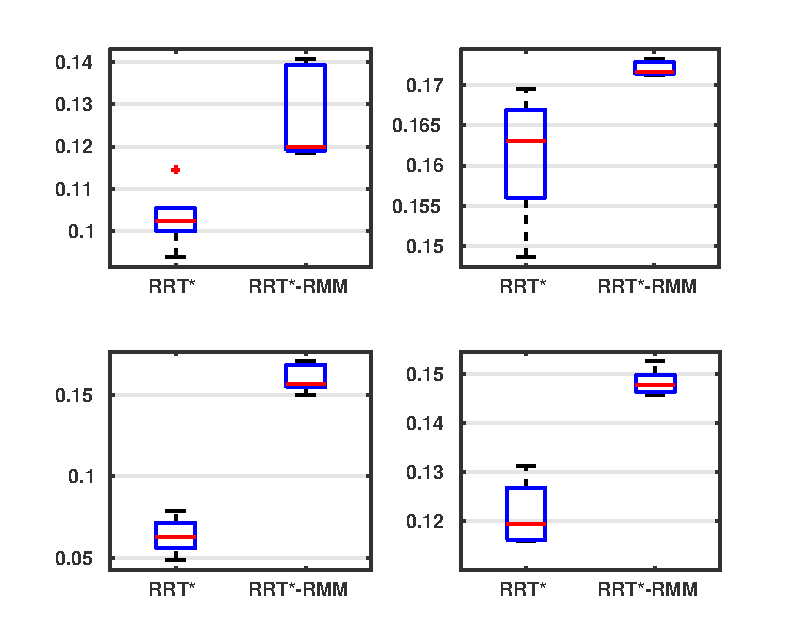}
  \caption{Box plot of the manipulability values obtained by RRT* and RRT*-RMM for all four objects. The order of these figures corresponds with the order of object figures shown in Fig.~\ref{Fig::objects}, \textit{i.e.}, (a) top left, (b) top right, (c) bottom left and (d) bottom right.}
  \label{Fig::boxplot}
  \vspace{0pt}
\end{figure}
%
The markers fix the initial and end point of the cutting path.  
These points can be provided by a human operator during real-world deployments. 
Fig.~\ref{fig::helmet_vrep} shows the point cloud of the helmet captured by the camera and visualised in V-REP.
The paths computed by RRT* and our proposed approach, RRT*-RMM, are shown with red and blue dotted lines, Fig.~\ref{fig::helmet_vrep}. 
These results show that RRT* and RRT*-RMM effectively generate cutting path on the object surface.  
Fig.~\ref{fig::helmet_man} also shows the manipulability corresponding to the paths obtained by RRT* and RRT*-RMM with red and blue lines, respectively.
\valcahnge{This figure shows although the paths obtained by RRT* and RRT*-RMM are very close, the path obtained by RRT*-RMM yield significantly higher capability of manipulation for the respective manipulator.}{This figure shows that our algorithm finds a path that has a significantly improved manipulability for the robot throughout the whole path.}
We see that RRT* generates a path specific just to the shape of the object. 

In contrast, RRT*-RMM computes paths not only specific to the shape of the object, but also specific to (i) the position of the object relative to the robot base frame and (ii) the kinematic chain of the robot.  
If we change either object position or the robotic arm, RRT*-RMM computes another path which is best for that scenario. 
As such, our algorithm always finds a path which is the best fit for the specific problem setting. 
Nonetheless, these changes can be easily embedded in the algorithm by using the URDF of the corresponding robot and the relative position of the object is captured by the RGB-D sensor calibrated in the robot base frame and the markers.

We performed similar experiments with all four objects shown in Fig.~\ref{Fig::objects}. 
Sample point clouds of 4 objects visualised in V-REP are shown in Fig.\ref{Fig::vrep_path} along with the computed path by RRT* and RRT*-RMM, red and blue respectively.  
In detail, Fig.~\ref{fig::cuboid_vrep} shows that the robot faces singularity if it follows the path obtained by RRT* for the flat object shown in Fig.~\ref{Fig::objects}c. 
In contrast, it does not experience this issue when using the path obtained with RRT*-RMM because the approach is explicitly designed to avoid such an issue. 
The paths visualised in V-REP in Fig.~\ref{Fig::vrep_path} correspond to the markers positions shown in Fig.~\ref{Fig::objects}.
These figures show that slight differences in the path obtained by RRT* and RRT*-RMM yield higher manipulation capability for the manipulator.
%
%

\begin{figure}[tb!]
    \includegraphics[trim=0cm 0cm 0cm 0cm, clip=true,scale=.39,angle = 0]{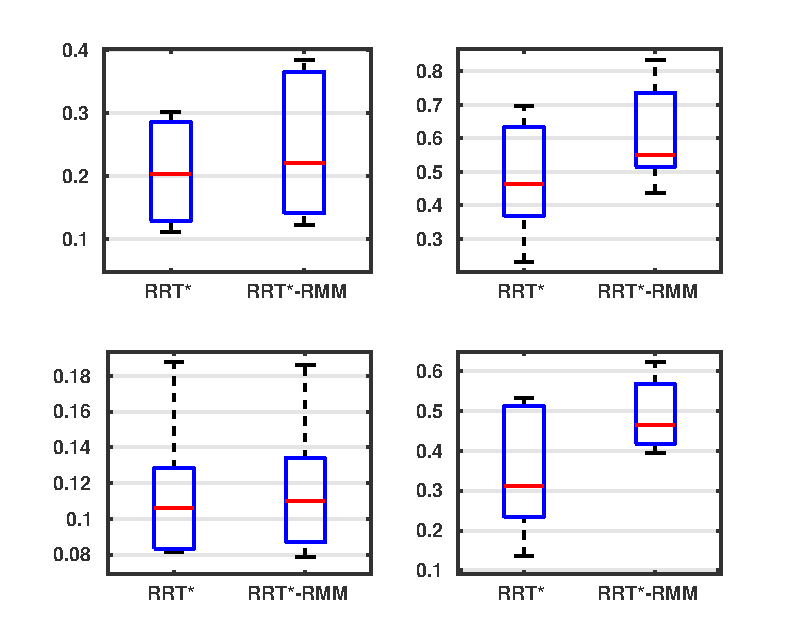}
  \vspace{-10pt}  
  \caption{This figure shows the path length found for the same objects. Every object has been tested using five goal positions, and the plots include the data for all the trials.}
  \label{fig::boxplot_dist}
  \vspace{-10pt}
\end{figure}

For every object, we repeated the experiment five times, each time with a different endpoint. 
We collected the data of manipulability and the length of the computed path by RRT* and RRT*-RMM. 
This allowed us to extensively measure how much our algorithm increases the path length, with respect to the RRT* baseline, and whether our approach has a beneficial effect on the manipulability. 
Fig.~\ref{Fig::boxplot} shows the box plot of obtained manipulability for all the objects. 
The results summarised in the box plot suggest that RRT*-RMM yields path with generally higher manipulability.
Because the nature of the approaches is random sample generations, the variation of the data captures the underlying behaviour of both approach. 
However, it is clear that the RRT*-RMM has generally improved the manipulability in the obtained path.  
RRT*-RMM also yields smaller variation of manipulability which is another desired characteristic. 
As we expected, the path length is increased with respect to the RRT* baseline, as it is traded-off for a higher manipulability throughout the path. 
Nonetheless, this increment is not very high as it is shown in Fig.~\ref{fig::boxplot_dist}. 
The increased path length is $\sim$10\% for the barrel, the curved object and the safety helmet, and $\sim$50\% for the flat object. 
%
\newline
This paper is accompanied by a video including experiments using Sawyer robot in V-REP simulation and using real Panda robot. 
\section{Conclusion}
This paper addresses the problem of constrained motion  planning from vision, which  enables a robot to move its   end-effector on an observed   surface, given start and destination points. We find robot trajectories which  maximise the robot’s  manipulability throughout the motion. This work has application in industrial problems of robotic rough cutting. 
Our approach uses a mapping between Euclidean space and Riemannian manifold to project the random samples generated by RRT* onto the object surface and vice versa.
This mapping step in our algorithm allows us to compute the path on an object surface using only a point cloud, without the need of a complete 3-D model. 
Moreover, we use a modified RRT* cost that sums a manipulability index and a cost based on the Euclidean distance between a new randomly generated point and the end point of the cut. 
The manipulability index added to the RRT* cost assures the generated samples of the path yield higher manipulability values. 
We presented a series of experiments with a Panda robot and a Sawyer robot. 
The experiments include computing the cutting paths on 4 different objects.
Since the core of RRT* and RRT*-RMM is random sample generation, we perfomed a statistical study that shows how RRT*-RMM improves the manipulability index while trading off on the length of the path, thus issuing longer paths with respect to the baseline of the classical RRT*.
%
%
%
%
While RRT*-RMM obtains an increased manipulation capability at the cost of increased path length, 
%
%
having longer paths, in our cutting problem of interest, which avoid robot-related issues, \textit{i.e.}, singularity, is acceptable. 

Future work includes the extension of this algorithm to non-Riemannian surfaces as this would allow the use of this algorithm to objects with sharp edges. 
Moreover, we are studying a suitable control architecture to include the proposed algorithm in a force control framework to enable effective cutting tasks and operations.
\bibliographystyle{ieeetr}
\bibliography{references}
\end{document}